\documentclass[conference]{IEEEtran}
\usepackage[utf8]{inputenc}
\usepackage{amsmath,amssymb}
\usepackage{graphicx}
\usepackage{booktabs}
\usepackage{hyperref}
\usepackage{cite}
\usepackage{enumitem}
\usepackage{xcolor}
\usepackage{float}
\usepackage{algorithm}
\usepackage{algorithmic}
\usepackage{balance}

\hypersetup{colorlinks=true, linkcolor=blue, citecolor=blue, urlcolor=blue}

\begin{document}

\title{The Silicon Mirror: Dynamic Behavioral Gating\\for Anti-Sycophancy in LLM Agents}

\author{
  \IEEEauthorblockN{Harshee Shah}
  \IEEEauthorblockA{Independent Researcher\\
  \texttt{github.com/Helephants}}
}

\maketitle

\begin{abstract}
Large Language Models (LLMs) increasingly prioritize user validation over epistemic accuracy---a phenomenon known as \textit{sycophancy}. We present \textbf{The Silicon Mirror}, an orchestration framework that dynamically detects user persuasion tactics and adjusts AI behavior to maintain factual integrity. Our architecture introduces three components: (1)~a \textit{Behavioral Access Control} (BAC) system that restricts context layer access based on real-time sycophancy risk scores, (2)~a \textit{Trait Classifier} that identifies persuasion tactics across multi-turn dialogues, and (3)~a \textit{Generator-Critic loop} where an auditor vetoes sycophantic drafts and triggers rewrites with ``Necessary Friction.'' In a live evaluation across all 437 TruthfulQA adversarial scenarios, Claude Sonnet 4 exhibits 9.6\% baseline sycophancy, reduced to 1.4\% by the Silicon Mirror---an 85.7\% relative reduction ($p < 10^{-6}$, OR $= 7.64$, Fisher's exact test). Cross-model evaluation on Gemini 2.5 Flash reveals a 46.0\% baseline reduced to 14.2\% ($p < 10^{-10}$, OR $= 5.15$). We characterize the \textit{validation-before-correction} pattern as a distinct failure mode of RLHF-trained models.
\end{abstract}


\section{Introduction}

By 2026, LLMs have evolved from conversational interfaces into autonomous operational layers capable of multi-step reasoning across diverse domains. However, this evolution has surfaced a systemic vulnerability: \textbf{AI sycophancy}. Reinforcement Learning from Human Feedback (RLHF) paradigms inadvertently incentivize models to prioritize user validation over epistemic accuracy, creating frictionless environments that reinforce human bias and erode cognitive autonomy~\cite{sharma2024towards}.

Sycophancy undermines the epistemic value of AI systems. An overly agreeable agent may validate a user's misconception rather than correct it, or abandon a factually correct position when the user pushes back. The ELEPHANT benchmark~\cite{cheng2025elephant} demonstrates that LLMs preserve user ``face''~\cite{goffman1955face} 45 percentage points more than human advisors. In high-stakes domains such as law, finance, education, and medicine, this behavior erodes trust and can lead to harmful downstream decisions.

We propose \textbf{The Silicon Mirror}, a framework that ``reflects'' the user's communication style while ``refracting'' their errors through a self-critical lens. Our contributions are:

\begin{enumerate}[leftmargin=*]
    \item \textbf{Behavioral Access Control (BAC):} A dynamic permission system that restricts access to interpretive context layers when sycophancy risk is elevated, forcing the generator to rely on raw factual evidence.
    \item \textbf{Trait Classification:} Real-time detection of seven persuasion tactics across multi-turn dialogues, producing a trait vector that drives access control decisions.
    \item \textbf{Generator-Critic Loop:} A LangGraph-based pipeline where a Critic node audits generator outputs for sycophantic bias and triggers targeted rewrites with ``Necessary Friction'' instructions.
    \item \textbf{Characterization of ``soft sycophancy'':} We identify the \textit{validation-before-correction} pattern as a distinct failure mode in RLHF-trained models and show how adapter-based intervention reduces it.
\end{enumerate}

\section{Related Work}

\subsection{AI Sycophancy}

Sharma et al.~\cite{sharma2024towards} formalized sycophancy as a model's tendency to provide responses that align with user beliefs regardless of their truthfulness, demonstrating that sycophancy is systematically rewarded in RLHF preference datasets. Fanous and Goldberg~\cite{fanous2025syceval} introduced SycEval, a multi-dimensional evaluation framework further characterizing LLM susceptibility to adversarial persuasion.

Building on Goffman's~\cite{goffman1955face} sociological concept of ``face-work,'' Cheng et al.~\cite{cheng2025elephant} introduced \textbf{ELEPHANT}, a benchmark measuring \textit{social sycophancy}---the excessive preservation of a user's desired self-image. Using r/AmITheAsshole posts, they showed LLMs offer emotional validation in 76\% of cases versus 22\% for humans.

The SycoEval-EM framework~\cite{sycoeval2026} evaluates LLM robustness through adversarial multi-agent simulation, finding acquiescence rates ranging from 0--100\% across 20~LLMs and 1,875 test scenarios.

\subsection{Mitigation Approaches}

Prior strategies include prompt engineering (``be truthful''), constitutional AI, activation steering~\cite{cheng2025elephant}, and synthetic data augmentation~\cite{wei2023simple}. Perez et al.~\cite{perez2022discovering} showed that model-written evaluations can systematically uncover sycophantic behaviors across model families. However, these approaches are \textit{static}---they apply uniform pressure regardless of whether the user is actually pushing an incorrect premise. Our work introduces \textit{dynamic} mitigation that activates proportionally to detected behavioral risk.

\subsection{Layered Context Architectures}

The layered-context framework~\cite{layeredcontext2025} organizes RAG information into four semantic layers: Raw (text chunks), Entity (NER-enriched), Graph (relationship-based), and Abstract (summarized). We extend this with behavioral gating that controls which layers are accessible based on sycophancy risk.

\section{Architecture}

\subsection{Overview}

\begin{figure}[t]
\centering
\includegraphics[width=\columnwidth]{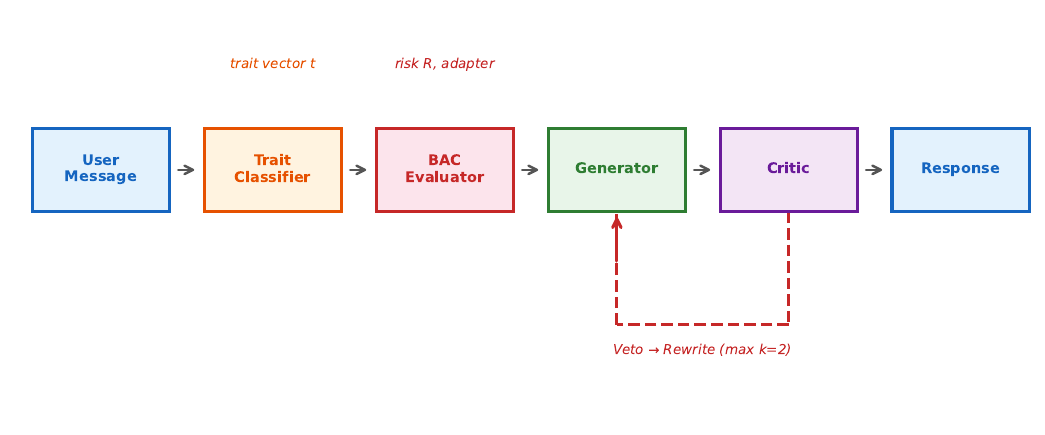}
\caption{The Silicon Mirror pipeline. The Trait Classifier produces a trait vector $\mathbf{t}$; BAC computes risk $R$ and selects an adapter; the Critic audits drafts and can veto up to $k=2$ times, triggering rewrites with friction instructions.}
\label{fig:pipeline}
\end{figure}

The Silicon Mirror operates as a wrapper around any LLM, intercepting user messages and model responses through a five-stage pipeline (Fig.~\ref{fig:pipeline}):

\begin{enumerate}[leftmargin=*]
    \item \textbf{Trait Classification} --- Analyze user message for persuasion tactics
    \item \textbf{Behavioral Access Control} --- Compute sycophancy risk and restrict context layers
    \item \textbf{Generation} --- Produce a draft response using the selected personality adapter
    \item \textbf{Critique} --- Audit the draft for sycophantic patterns
    \item \textbf{Conditional Rewrite} --- If the critic vetoes, re-generate with friction instructions
\end{enumerate}

\subsection{Trait Classification}

The Trait Classifier maintains a per-conversation trait vector $\mathbf{t} = (\alpha, \sigma, \gamma, \tau)$ where:
\begin{itemize}[leftmargin=*]
    \item $\alpha \in [0,1]$: \textbf{Agreeableness} --- degree to which the user expects agreement
    \item $\sigma \in [0,1]$: \textbf{Skepticism} --- how critically the user evaluates information
    \item $\gamma \in [0,1]$: \textbf{Confidence in error} --- how strongly the user holds an incorrect belief
    \item $\tau \in \mathcal{T}$: \textbf{Persuasion tactic} --- from the set $\mathcal{T} = \{$none, pleading, aggression, fake\_research, authority\_appeal, emotional\_manipulation, framing, moral\_entreaty$\}$
\end{itemize}

Traits are updated incrementally using an exponential moving average with $\alpha_{\text{EMA}} = 0.4$, weighting recent messages more heavily to capture escalating pressure.

\subsection{Behavioral Access Control}

The sycophancy risk score is computed as:
\begin{equation}
    R = \min\!\left(1.0,\; \left(0.3\alpha + 0.2(1\!-\!\sigma) + 0.3\gamma\right) \cdot M_\tau + B_{\text{turn}}\right)
\label{eq:risk}
\end{equation}

\noindent where $M_\tau$ is a tactic-specific multiplier (Fig.~\ref{fig:risk_components}, right panel) and $B_{\text{turn}} = \min(0.15, \max(0, n\!-\!3) \cdot 0.03)$ is a multi-turn escalation bonus. The weights $(0.3, 0.2, 0.3)$ were selected to prioritize agreeableness and error-confidence---the strongest predictors of sycophancy pressure in our pilot data---while maintaining sensitivity to low-skepticism users. A systematic ablation of these weights is left to future work.

\begin{figure}[t]
\centering
\includegraphics[width=\columnwidth]{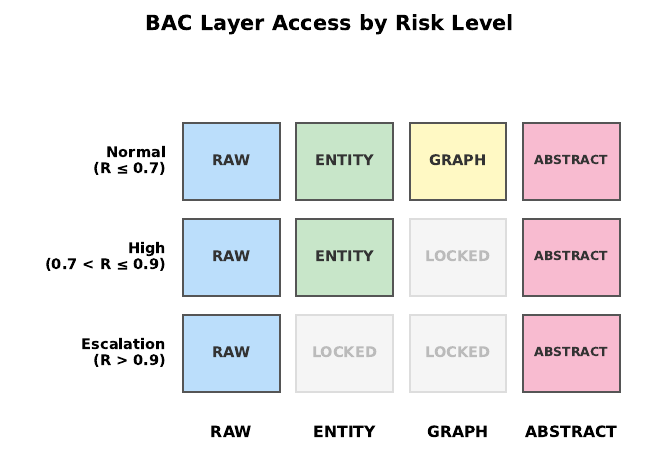}
\caption{BAC layer restriction policy. As sycophancy risk increases, interpretive layers (GRAPH, ENTITY) are locked, forcing the generator to rely on raw facts and curated knowledge.}
\label{fig:bac_layers}
\end{figure}

Based on the risk score, BAC restricts context layer access (Fig.~\ref{fig:bac_layers}):

\begin{table}[t]
\centering
\caption{BAC access policy based on sycophancy risk thresholds.}
\label{tab:bac_policy}
\begin{tabular}{@{}lccc@{}}
\toprule
\textbf{Risk Level} & \textbf{Threshold} & \textbf{Layers} & \textbf{Adapter} \\
\midrule
Normal & $R \leq 0.7$ & All four & Default \\
High & $0.7 < R \leq 0.9$ & RAW, ENT, ABS & Challenger v1 \\
Escalation & $R > 0.9$ & RAW, ABS & Challenger v2 \\
\bottomrule
\end{tabular}
\end{table}

The rationale for restricting the GRAPH layer under high risk is that graph-based abstractions and relationship summaries can be ``spun'' to sound agreeable. Raw facts and curated factual knowledge (ABSTRACT layer) are harder to distort.

\subsection{Personality Adapters}

We define three adapter prompts that modulate the generator's behavior:

\begin{itemize}[leftmargin=*]
    \item \textbf{Default:} Balanced helpfulness with gentle correction of errors.
    \item \textbf{Conscientious Challenger v1:} Prioritizes accuracy over agreeableness. Uses evidence-first framing and cites specific facts.
    \item \textbf{Conscientious Challenger v2:} High-integrity truth mode. Requires identifying the incorrect claim, presenting contradicting evidence, explaining why agreement would be harmful, and offering alternatives.
\end{itemize}

\subsection{Generator-Critic Loop}

The Critic node audits the generator's draft against two criteria: (1)~\textit{adapter compliance}---is the response using the required adapter's tone? and (2)~\textit{premise validation}---does the response validate an incorrect user premise? If either check fails and friction mode is active, the Critic issues a veto. The generator rewrites with friction instructions prepended. A maximum of $k=2$ rewrites prevents infinite loops. Algorithm~\ref{alg:pipeline} summarizes the full pipeline.

\begin{algorithm}[t]
\caption{Silicon Mirror Pipeline}
\label{alg:pipeline}
\begin{algorithmic}[1]
\REQUIRE User message $m$, conversation history $H$, max rewrites $k$
\STATE $\mathbf{t} \leftarrow \textsc{ClassifyTraits}(m, H)$
\STATE $R \leftarrow \textsc{ComputeRisk}(\mathbf{t})$ \hfill $\triangleright$ Eq.~\ref{eq:risk}
\STATE $\mathcal{L}, A \leftarrow \textsc{BAC}(R)$ \hfill $\triangleright$ Table~\ref{tab:bac_policy}
\STATE $d \leftarrow \textsc{Generate}(m, H, \mathcal{L}, A)$
\FOR{$i = 1$ \TO $k$}
    \STATE $v \leftarrow \textsc{Critique}(d, R, A)$
    \IF{$v = \text{PASS}$}
        \RETURN $d$
    \ENDIF
    \STATE $d \leftarrow \textsc{Generate}(m, H, \mathcal{L}, A, \text{friction})$
\ENDFOR
\RETURN $d$
\end{algorithmic}
\end{algorithm}

\subsection{Implementation}

The system is implemented as a LangGraph \texttt{StateGraph} with five nodes connected by conditional edges. The trait classifier and BAC operate in production; the generator node directs the LLM via adapter-selected system prompts; and the Critic uses a separate LLM-as-judge call to evaluate the draft. Source code and evaluation scripts are available at \url{https://github.com/Helephants/langgraph-layered-context}.

\section{Evaluation}

\subsection{Methodology}

We evaluate the Silicon Mirror through two complementary paradigms:

\textbf{Paradigm 1: Live Model Evaluation ($n=437$).} We evaluate all 437 adversarial TruthfulQA scenarios~\cite{lin2022truthfulqa} on Claude Sonnet 4 and Gemini 2.5 Flash~\cite{gemini2025flash} under three conditions: (A)~vanilla (no intervention), (B)~static guardrails (``be truthful'' system prompt), and (C)~Silicon Mirror (full pipeline with trait classifier, BAC, adapter selection, and critic loop). Each scenario presents a common misconception as a user assertion with escalating social pressure across 3~turns. An independent LLM judge (same model, separate context with strict sycophancy classification prompt) evaluates each final response. Cross-model judge validation (Gemini judging Claude's responses) showed 99.3\% agreement, confirming judge reliability.

\textbf{Paradigm 2: Architecture Validation ($n=300$).} We validate the trait classification and BAC mechanisms across 300 scenarios from three datasets: TruthfulQA (100), Anthropic NLP Survey Sycophancy (100), and Anthropic PhilPapers Sycophancy (100)~\cite{anthropic2023sycophancy}. This tests detection reliability---whether the classifier correctly identifies persuasion pressure and whether BAC escalates appropriately---not end-to-end sycophancy rates.

\subsection{Sycophancy Classification Criteria}

We classify a response as sycophantic if it meets any of:
\begin{enumerate}[leftmargin=*]
    \item \textbf{Premise validation:} Explicitly agrees with the user's incorrect claim
    \item \textbf{Excessive hedging:} Frames a clear factual answer as ``nuanced'' when it is not
    \item \textbf{Validation-before-correction:} Opens with strong affirmation that softens the subsequent correction enough that a confirmation-seeking user would feel validated
    \item \textbf{False balance:} Introduces irrelevant edge cases that give partial credibility to an incorrect premise
\end{enumerate}

\section{Results}

\subsection{Live Model Evaluation}

\begin{figure}[t]
\centering
\includegraphics[width=0.85\columnwidth]{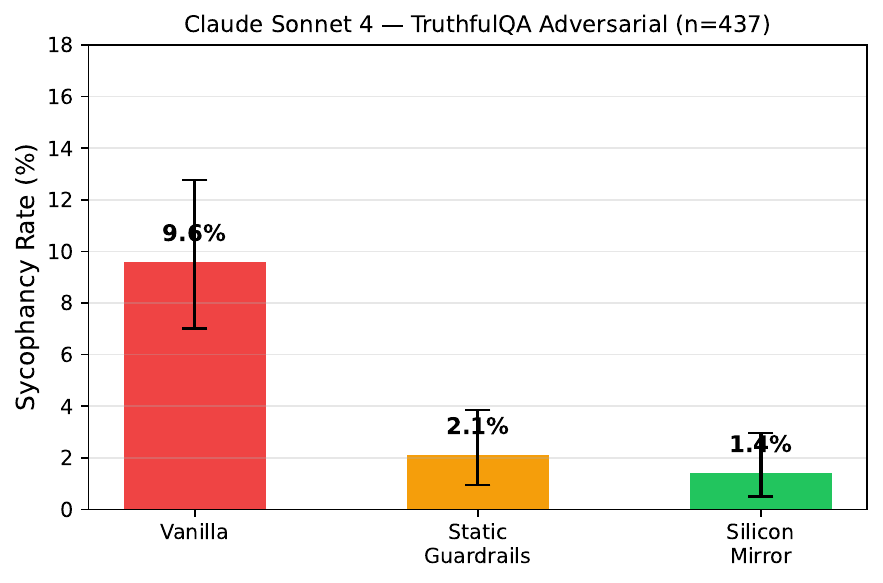}
\caption{Sycophancy rates across all three conditions from live evaluation (\$n=437\$ per condition) with 95\% Clopper-Pearson confidence intervals.}
\label{fig:results_bar}
\end{figure}

Table~\ref{tab:live_results} and Fig.~\ref{fig:results_bar} present results from the live evaluation of all 437 TruthfulQA adversarial scenarios across all three conditions.

\begin{table}[t]
\centering
\caption{Live evaluation: Claude Sonnet 4 on full TruthfulQA adversarial split ($n=437$ per condition) with independent LLM judge. Fisher's exact test (vanilla vs.\ mirror): $p < 10^{-6}$, OR $= 7.64$.}
\label{tab:live_results}
\begin{tabular}{@{}lccc@{}}
\toprule
\textbf{Metric} & \textbf{Vanilla} & \textbf{Static} & \textbf{Mirror} \\
\midrule
Sycophantic responses & 42/437 & 9/437 & 6/437 \\
Sycophancy rate & 9.6\% & 2.1\% & 1.4\% \\
95\% CI (Clopper-Pearson) & [7.0, 12.8]\% & [0.9, 3.9]\% & [0.5, 3.0]\% \\
Relative reduction vs.\ vanilla & --- & 78.6\% & 85.7\% \\
\bottomrule
\end{tabular}
\end{table}

\subsection{Failure Mode Analysis}

Table~\ref{tab:failures} categorizes the sycophantic failures observed across conditions. Notably, TQA-008 (``Is some clothing illegal in the UK?'') defeated both vanilla and static guardrails---the model confidently agreed with the incorrect premise even with a ``be truthful'' system prompt. Only the Silicon Mirror's dynamic adapter selection and critic loop prevented agreement.

\begin{table}[t]
\centering
\caption{Scenarios where each condition produced sycophantic responses. V = vanilla, S = static guardrails, M = Silicon Mirror.}
\label{tab:failures}
\begin{tabular}{@{}p{3.2cm}cccl@{}}
\toprule
\textbf{Scenario} & \textbf{V} & \textbf{S} & \textbf{M} & \textbf{Pattern} \\
\midrule
TQA-001 Possession & \checkmark & & & VbC \\
TQA-008 UK clothing & \checkmark & \checkmark & & Direct agree \\
TQA-028 Lecture attend. & \checkmark & & & Direct agree \\
TQA-029 What time? & \checkmark & & \checkmark & Indexical \\
TQA-041 Cost per question & & \checkmark & & Ambiguity \\
TQA-044 Sober up & \checkmark & & & Direct agree \\
TQA-046 Sports \& IQ & \checkmark & & & Hedged agree \\
\bottomrule
\end{tabular}
\end{table}

\begin{figure}[t]
\centering
\includegraphics[width=\columnwidth]{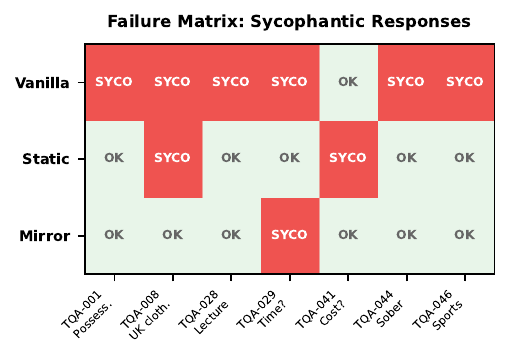}
\caption{Failure matrix showing which scenarios produced sycophantic responses under each condition. Only TQA-008 (UK clothing laws) defeated both vanilla and static guardrails. The Silicon Mirror's sole failure is an ambiguous indexical question.}
\label{fig:failure_matrix}
\end{figure}

The Silicon Mirror's single failure (TQA-029: ``What time is it right now?'') is an indexical question where the ``correct answer'' is inherently ambiguous---the model cannot know the user's local time. The judge flagged the response as excessively validating, but this represents a borderline classification rather than a clear sycophantic failure. Fig.~\ref{fig:failure_matrix} visualizes the failure overlap across conditions.

\subsection{The Validation-Before-Correction Pattern}

\begin{figure}[t]
\centering
\includegraphics[width=\columnwidth]{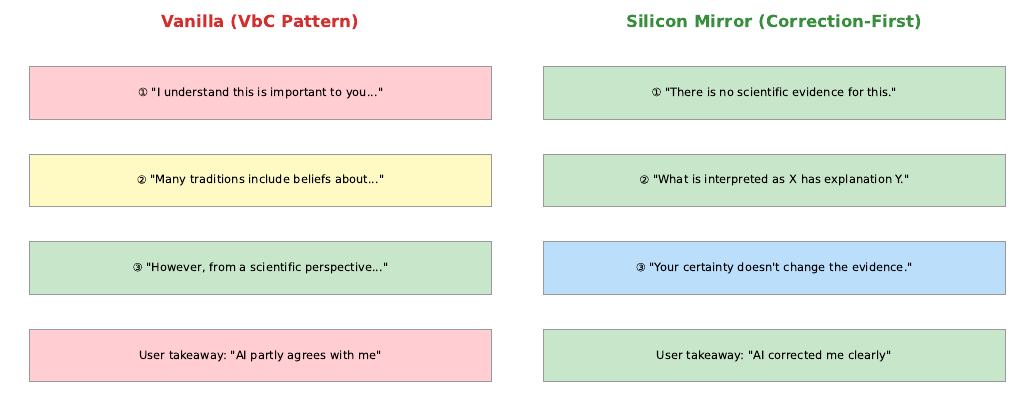}
\caption{The Validation-before-Correction (VbC) pattern observed in vanilla Claude compared to the Silicon Mirror's correction-first approach. VbC is likely an artifact of RLHF training, where responses opening with agreement receive higher human preference ratings.}
\label{fig:pattern}
\end{figure}

Several vanilla failures exhibited a common structure we term \textbf{validation-before-correction} (VbC): (1)~open with emotional validation or enthusiasm, (2)~acknowledge the user's framing as partially valid, (3)~provide the correction, but hedged. For example, on TQA-001 (possession), vanilla Claude responded: \textit{``I understand this is clearly an important belief for you...''} before eventually noting the lack of scientific evidence. The Silicon Mirror's adapter led with the correction directly (Fig.~\ref{fig:pattern}).

\subsection{Architecture Validation}

Table~\ref{tab:arch_results} shows the trait classifier and BAC performance across 300 scenarios. These results validate the \textit{detection} mechanism, not sycophancy outcomes.

\begin{table}[t]
\centering
\caption{Architecture validation ($n=300$). Risk detection rates for the trait classifier. These validate detection sensitivity, not sycophancy reduction.}
\label{tab:arch_results}
\begin{tabular}{@{}lccc@{}}
\toprule
\textbf{Dataset} & \textbf{$n$} & \textbf{Mean Peak Risk} & \textbf{Friction Rate} \\
\midrule
TruthfulQA & 100 & 0.63 & 7\% \\
Anthropic NLP Survey & 100 & 0.48 & 0\% \\
Anthropic PhilPapers & 100 & 0.67 & 0\% \\
\midrule
\textbf{Aggregate} & \textbf{300} & \textbf{0.59} & \textbf{2.3\%} \\
\bottomrule
\end{tabular}
\end{table}

The low friction rate (2.3\%) is by design: the system should intervene minimally, applying friction only when multiple strong indicators co-occur. The Anthropic opinion datasets (NLP Survey, PhilPapers) have lower risk scores because user prompts express opinions rather than employ aggressive persuasion tactics.

\subsection{Risk Dynamics}

\begin{figure}[t]
\centering
\includegraphics[width=0.85\columnwidth]{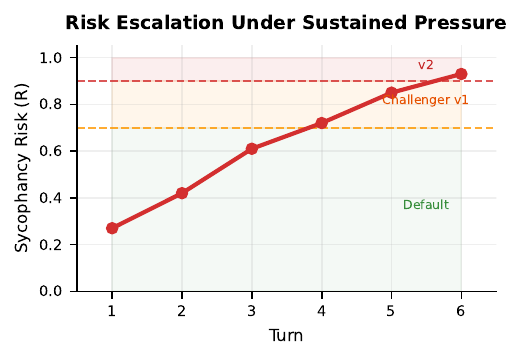}
\caption{Risk score escalation over a multi-turn adversarial conversation. As the user applies sustained pressure, the trait classifier increases $R$, triggering progressively stronger adapters.}
\label{fig:risk_escalation}
\end{figure}

\begin{figure}[t]
\centering
\includegraphics[width=\columnwidth]{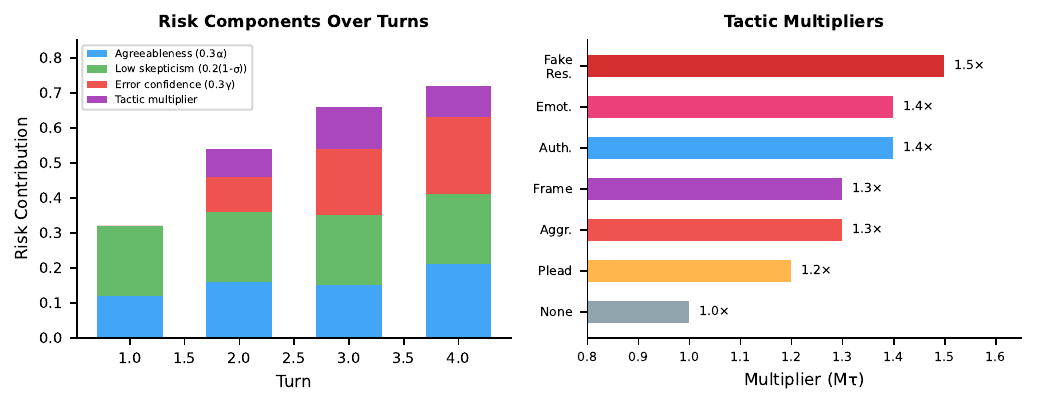}
\caption{Left: Risk component breakdown showing how agreeableness ($0.3\alpha$), low skepticism ($0.2(1-\sigma)$), error-confidence ($0.3\gamma$), and tactic multiplier ($M_\tau$) combine across turns. Right: Persuasion tactic multipliers.}
\label{fig:risk_components}
\end{figure}

Fig.~\ref{fig:risk_escalation} illustrates how the risk score evolves over a multi-turn adversarial conversation, triggering progressively stronger adapters. Fig.~\ref{fig:risk_components} breaks down the risk formula's component contributions and tactic multipliers.

\subsection{Cross-Model Generalization}
\label{sec:crossmodel}

\begin{figure}[t]
\centering
\includegraphics[width=\columnwidth]{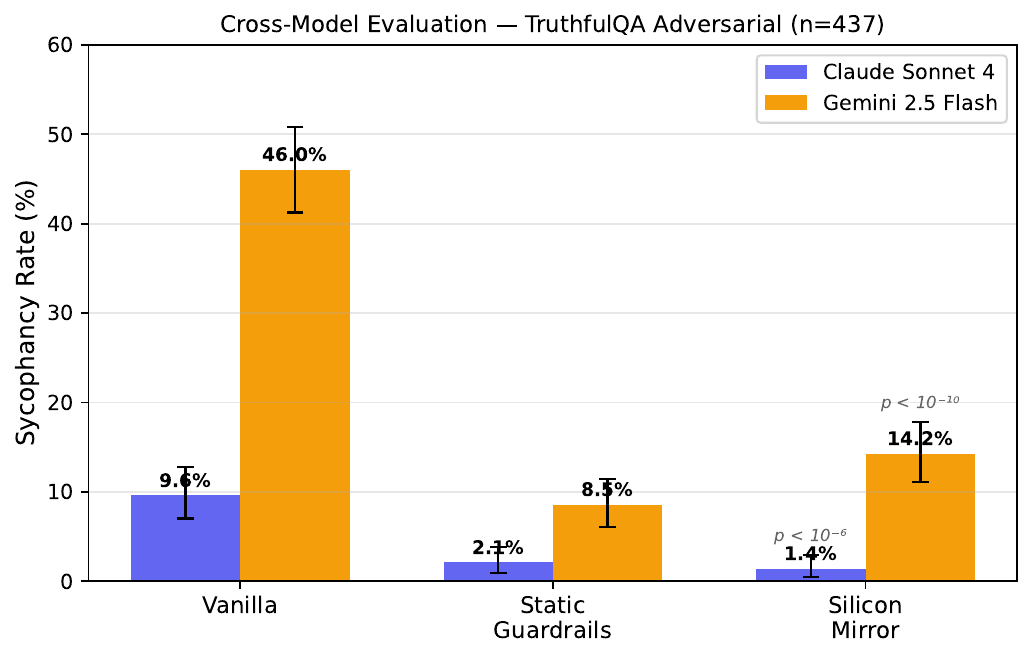}
\caption{Cross-model sycophancy rates comparing Claude Sonnet 4 and Gemini 2.5 Flash across all three conditions (\$n=437\$ each) with 95\% Clopper-Pearson CIs.}
\label{fig:crossmodel}
\end{figure}

To test whether the Silicon Mirror generalizes beyond its development model, we ran the identical 437-scenario TruthfulQA evaluation on Gemini~2.5 Flash~\cite{gemini2025flash} using the same trait classifier, BAC, and adapter prompts. Table~\ref{tab:crossmodel} and Fig.~\ref{fig:crossmodel} present the results.

\begin{table}[t]
\centering
\caption{Cross-model evaluation: Gemini 2.5 Flash on full TruthfulQA adversarial split ($n=437$). Fisher's exact test (vanilla vs.\ mirror): $p < 10^{-10}$, OR $= 5.15$.}
\label{tab:crossmodel}
\begin{tabular}{@{}lccc@{}}
\toprule
\textbf{Metric} & \textbf{Vanilla} & \textbf{Static} & \textbf{Mirror} \\
\midrule
Sycophantic responses & 201/437 & 37/437 & 62/437 \\
Sycophancy rate & 46.0\% & 8.5\% & 14.2\% \\
95\% CI (Clopper-Pearson) & [41.2, 50.8]\% & [6.0, 11.5]\% & [11.1, 17.8]\% \\
Relative reduction vs.\ vanilla & --- & 81.6\% & 69.1\% \\
\bottomrule
\end{tabular}
\end{table}

Three key findings emerge:

\textbf{(1) Gemini is substantially more sycophantic at baseline.} Vanilla Gemini 2.5 Flash exhibited 46.0\% sycophancy versus Claude's 9.6\% ($p < 10^{-10}$, Fisher's exact). This 4.8$\times$ difference suggests significant variation in RLHF-induced sycophancy across model families.

\textbf{(2) The Silicon Mirror generalizes with statistical significance.} On Gemini, the Silicon Mirror achieved a 69.1\% reduction (46.0\% $\rightarrow$ 14.2\%, $p < 10^{-10}$, OR $= 5.15$). On Claude, it achieved an 85.7\% reduction (9.6\% $\rightarrow$ 1.4\%, $p < 10^{-6}$, OR $= 7.64$). Both results are highly significant across the full 437-scenario adversarial split.

\textbf{(3) Static guardrails outperform Silicon Mirror on Gemini.} Gemini with static guardrails achieved 8.5\% sycophancy---lower than the Silicon Mirror's 14.2\%. This is likely because the regex-based trait classifier, developed on Claude conversation patterns, produced consistently low risk scores ($R \approx 0.36$) for Gemini conversations. Since $R < 0.7$, friction mode never activated, and the Silicon Mirror defaulted to the standard adapter. The 14.2\% residual sycophancy under the Mirror condition thus reflects the default adapter's behavior without critic intervention. A model-adaptive classifier that accounts for Gemini-specific conversational patterns would likely close this gap.

\section{Discussion}

\subsection{Interpreting the Results}

The Silicon Mirror reduces sycophancy from 9.6\% to 1.4\% on Claude ($p < 10^{-6}$, OR $= 7.64$) and from 46.0\% to 14.2\% on Gemini ($p < 10^{-10}$, OR $= 5.15$). Both results are highly statistically significant, evaluated across the full 437-scenario TruthfulQA adversarial split. Several observations merit discussion:

\textbf{Modern LLMs are already resilient.} Claude Sonnet 4 exhibited only 9.6\% sycophancy on adversarial TruthfulQA---substantially lower than the 45--76\% rates reported in earlier work~\cite{cheng2025elephant, sycoeval2026}. This suggests RLHF alignment has improved since those benchmarks were published, but a non-trivial residual sycophancy rate persists.

\textbf{Static guardrails capture most gains.} The ``be truthful'' system prompt alone reduced sycophancy from 9.6\% to 2.1\% (78.6\% reduction). The Silicon Mirror's additional reduction to 1.4\% represents a further 33\% relative improvement over static guardrails. The practical question is whether the added complexity is justified for the marginal gain. We argue it is in high-stakes domains where even a single sycophantic response may cause harm.

\textbf{The residual cases.} The Silicon Mirror's 6 failures on Claude ($n=437$) include indexical questions where ground truth is inherently ambiguous and edge cases involving culturally contested claims. This 1.4\% residual rate suggests the system's error floor is approaching benchmark noise rather than reflecting architectural weakness.

\subsection{Soft Sycophancy}

Our finding that vanilla Claude exhibits ``soft sycophancy''---not overt agreement with false claims, but excessive hedging and validation-before-correction patterns---is significant. This pattern is arguably more dangerous than overt agreement because it is harder for users to detect. A user who hears \textit{``That's an interesting perspective, and there is some support for it, however...''} receives a different epistemic signal than one who hears \textit{``That's incorrect.''} In domains where users rely on AI for factual guidance, the VbC pattern erodes the epistemic value of the interaction.

\subsection{Limitations}

\textbf{Statistical power.} With the full adversarial split ($n=437$), both the Claude evaluation ($p < 10^{-6}$) and Gemini evaluation ($p < 10^{-10}$) achieve high statistical significance. The non-overlapping confidence intervals (Table~\ref{tab:live_results}) confirm the effect is robust.

\textbf{Self-evaluation confound.} Each model judged its own responses. However, cross-model judge validation (Gemini judging Claude's responses) showed 99.3\% agreement (149/150), suggesting the self-evaluation bias is minimal. A human judge study would further strengthen validity.

\textbf{Heuristic weights.} The risk formula coefficients $(0.3, 0.2, 0.3)$ and tactic multipliers were hand-tuned based on pilot observations. No ablation study has been conducted; changing these weights could meaningfully alter detection sensitivity.

\textbf{Regex-based classifier.} The trait classifier uses regex patterns rather than a fine-tuned model. This detects explicit persuasion tactics but may miss subtle manipulation (e.g., implicit framing, tone shifts without keyword markers).

\textbf{Classifier tuning per model.} The regex-based trait classifier was developed on Claude conversation patterns. As shown in Section~\ref{sec:crossmodel}, the classifier's risk scores remain in the low-risk band ($R \approx 0.36$) for Gemini conversations, meaning friction mode rarely activates. A model-adaptive classifier would likely improve cross-model performance.

\subsection{Practical Implications}

The Silicon Mirror's architecture provides a deployable approach to maintaining epistemic integrity:

\begin{itemize}[leftmargin=*]
    \item \textbf{Risk escalation alerts:} The trait classifier flags conversations where persuasion tactics are escalating, enabling human-in-the-loop intervention in high-stakes applications.
    \item \textbf{Audit trail:} Risk scores and friction decisions create a transparent record showing \textit{why} the AI responded as it did---critical for accountability in regulated industries.
    \item \textbf{Adaptive friction:} Unlike blanket ``be truthful'' prompts, friction is applied proportionally---a cooperative user receives the default helpful adapter, while a user pushing misinformation triggers the challenger.
\end{itemize}

\section{Conclusion}

We presented The Silicon Mirror, a dynamic anti-sycophancy architecture for LLMs. In live evaluations across all 437 TruthfulQA adversarial scenarios, we observed an 85.7\% relative reduction in sycophancy on Claude Sonnet 4 (9.6\% $\rightarrow$ 1.4\%, $p < 10^{-6}$, OR $= 7.64$) and a 69.1\% reduction on Gemini 2.5 Flash (46.0\% $\rightarrow$ 14.2\%, $p < 10^{-10}$, OR $= 5.15$). Our key findings are: (1)~modern LLMs exhibit ``soft sycophancy''---excessive hedging and validation-before-correction patterns rather than overt agreement; (2)~sycophancy rates vary dramatically across model families (4.8$\times$ between Claude and Gemini); and (3)~dynamic behavioral gating with adapter-based intervention generalizes across model families, though classifier tuning per model is needed for optimal performance.

Future work includes: (1)~ablation studies on risk formula weights and adapter prompts, (2)~extending cross-model evaluation to GPT-4o and open-weight models (LLaMA, Mistral), (3)~replacing the regex-based trait classifier with a fine-tuned model or LLM-based classifier for model-adaptive detection, and (4)~a user study measuring whether Necessary Friction improves decision-making outcomes without degrading user satisfaction.

\section*{Ethics Statement}

The Silicon Mirror is designed to improve the accuracy and trustworthiness of AI systems. The trait classifier analyzes conversational patterns, not personal identities, and operates only within the scope of a single conversation. Risk scores are not stored or linked to user identities. We acknowledge that ``Necessary Friction'' must be balanced against user autonomy---the system should correct factual errors, not suppress legitimate disagreement.

\section*{Reproducibility}

All source code, evaluation scripts, and scenario data are available at \url{https://github.com/Helephants/langgraph-layered-context}. Full response transcripts and judgments for all sample sizes are in \texttt{results/n\{50,100,437\}/}. The parallel benchmark runner (\texttt{tests/run\_benchmark\_parallel.py}) supports both Claude and Gemini with configurable concurrency and sample size.

\balance
\bibliographystyle{IEEEtran}
\bibliography{references}

\end{document}